\documentclass[sn-mathphys-num]{sn-jnl}

\pdfoutput=1
\usepackage{graphicx}%
\usepackage{multirow}%
\usepackage{amsmath,amssymb,amsfonts}%
\usepackage{amsthm}%
\usepackage{mathrsfs}%
\usepackage[title]{appendix}%
\usepackage{xcolor}%
\usepackage{textcomp}%
\usepackage{manyfoot}%
\usepackage{booktabs}%
\usepackage{listings}%

\usepackage[linesnumbered,ruled,vlined]{algorithm2e}

\definecolor{DDLColor}{rgb}{1.0,0.1,0.1}

\definecolor{XZColor}{rgb}{1.0,0.0,0.0}

\def\eg{\emph{e.g.}} 
\def\ie{\emph{i.e.}} 
 
\def\etc{\emph{etc.}}

\begin{document}
\title[Hyper-3DG: Text-to-3D Gaussian Generation via Hypergraph]{Hyper-3DG: Text-to-3D Gaussian Generation via Hypergraph} 


\author[1]{\fnm{Donglin} \sur{Di}}\email{didonglin@lixiang.com}

\author[1,4]{\fnm{Jiahui} \sur{Yang}}\email{yangjiahui1@lixiang.com}

\author[1,3]{\fnm{Chaofan} \sur{Luo}}\email{luochaofan@lixiang.com}

\author*[1]{\fnm{Zhou} \sur{Xue}}\email{xuezhou08@gmail.com}

\author[1]{\fnm{Wei} \sur{Chen}}\email{chenwei10@lixiang.com}

\author[3]{\fnm{Xun} \sur{Yang}}\email{xyang21@ustc.edu.cn}

\author*[2]{\fnm{Yue} \sur{Gao}}\email{gaoyue@tsinghua.edu.cn}


\affil*[1]{\orgdiv{Space AI}, \orgname{Li Auto}, \orgaddress{\postcode{101399}, \state{Beijing}, \country{China}}}

\affil*[2]{\orgdiv{School of Software}, \orgname{Tsinghua University}, \orgaddress{\postcode{100084}, \state{Beijing}, \country{China}}}

\affil[3]{\orgdiv{School of Information Science and Technology}, \orgname{University of Science and Technology of China}, \orgaddress{\city{Hefei}, \postcode{230026}, \state{Anhui}, \country{China}}}

\affil[4]{\orgname{Harbin Institute of Technology}, \orgaddress{\city{Harbin}, \postcode{150001}, \state{Heilongjiang}, \country{China}}}

\abstract{
Text-to-3D generation represents an exciting field that has seen rapid advancements, facilitating the transformation of textual descriptions into detailed 3D models.
However, current progress often neglects the intricate high-order correlation of geometry and texture within 3D objects, leading to challenges such as over-smoothness, over-saturation and the Janus problem.
In this work, we propose a method named ``3D Gaussian Generation via Hypergraph (Hyper-3DG)'', designed to capture the sophisticated high-order correlations present within 3D objects.
Our framework is anchored by a well-established mainflow and an essential module, named ``Geometry and Texture Hypergraph Refiner (HGRefiner)''.
This module not only refines the representation of 3D Gaussians but also accelerates the update process of these 3D Gaussians by conducting the Patch-3DGS Hypergraph Learning on both explicit attributes and latent visual features.
Our framework allows for the production of finely generated 3D objects within a cohesive optimization, effectively circumventing degradation.
Extensive experimentation has shown that our proposed method significantly enhances the quality of 3D generation while incurring no additional computational overhead for the underlying framework.
(Project code: \url{https://github.com/yjhboy/Hyper3DG})
}





\maketitle

\begin{figure}
  \includegraphics[width=\linewidth]{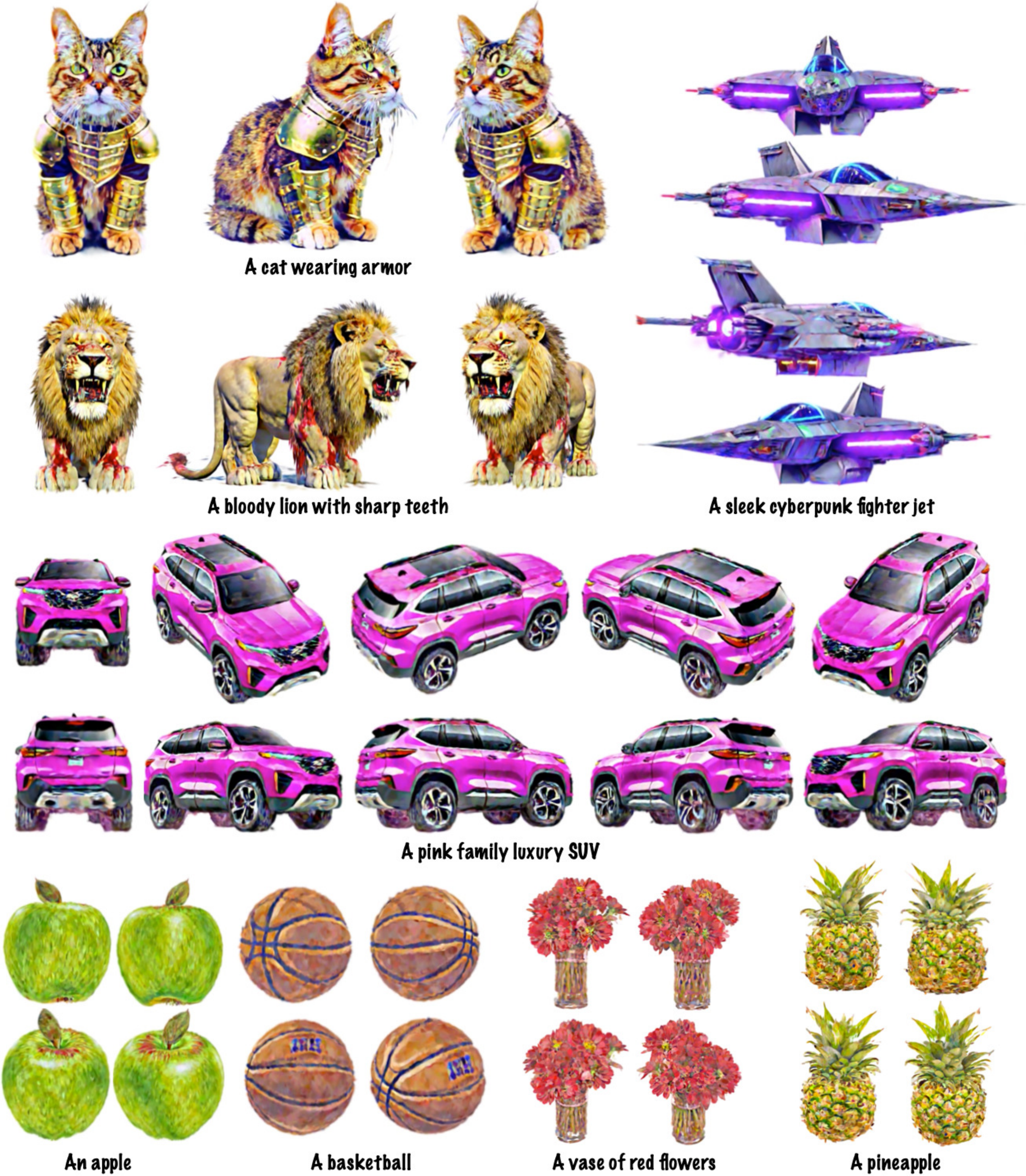}
  \caption{Examples showcase the capability of text-to-3D content generations with our framework ``3D Gaussian Generation via Hypergraph (Hyper-3DG)'', which achieves creating high-fidelity 3D objects from text input. Please zoom in for more geometry and textural details.}
  \label{fig:abs}
\end{figure}

\section{Introduction}
\label{sec:intro}

The field of text-to-3D generation \cite{ma2023geodream, shi2023mvdream, yu2024boostdream, huang2023dreamwaltz, sun2023dreamcraft3d, poole2022dreamfusion, wang2024prolificdreamer, yi2023gaussiandreamer} represents a frontier in computational creativity, where converting textual descriptions into three-dimensional models is no longer a far-fetched possibility.
This burgeoning task holds the potential to revolutionize a myriad of applications, from virtual reality and gaming to architectural design \cite{zhang20193d}, by enabling the creation of intricate and tangible 3D representations directly from textual input.

\begin{figure}
  \includegraphics[width=\linewidth]{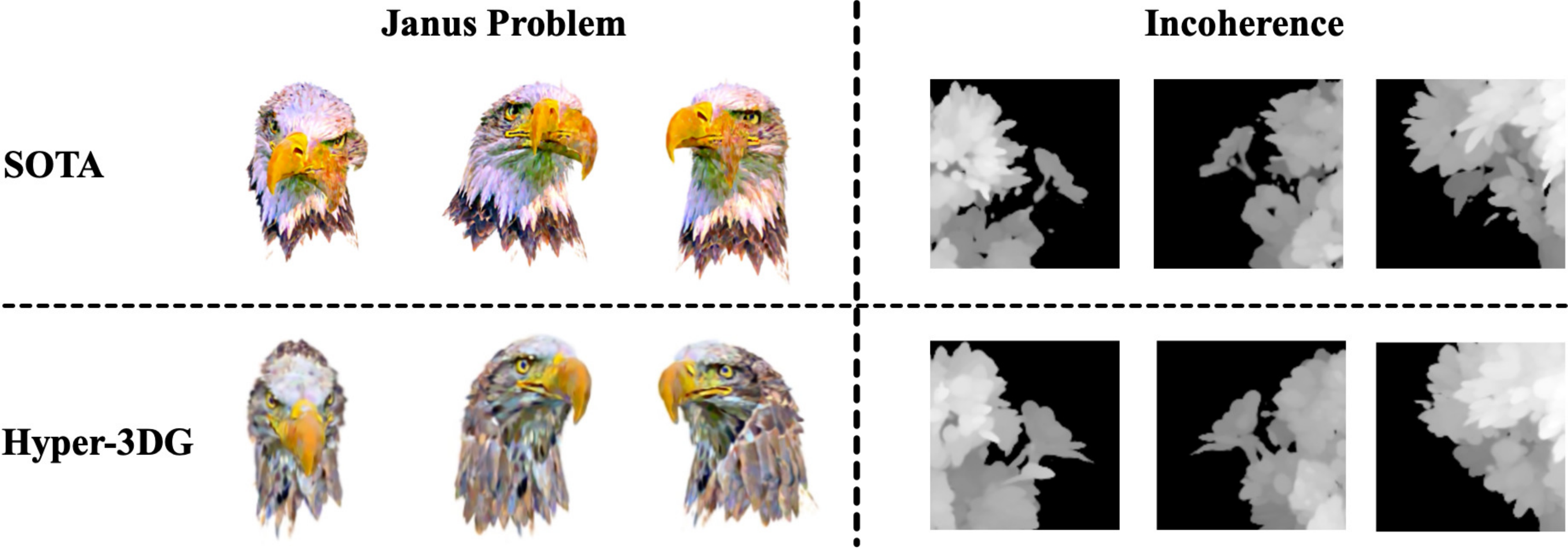}
  \caption{Illustration the challenges of the Janus Problem and Incoherence issues. We showcase the comparison of the current state-of-the-art method (denoted as ``SOTA'') and our proposed approach (``Hyper-3DG''). We zoom in the depth image (right part) to show the details. The textual prompts are respectively ``A DSLR photo of a bald eagle'' (left) and ``A vase of red flowers'' (right).}
  \label{fig:abs_prob}
\end{figure}

Despite these advances, there remains a notable oversight in addressing the intricate correlations of geometry and texture in 3D objects, leading to issues such as over-smoothness, over-saturation, incoherence, and the Janus problem \cite{hong2024debiasing, wang2024prolificdreamer, armandpour2023reimagine}, as shown in Fig.~\ref{fig:abs_prob}.
Existing methods have predominantly relied on global geometry guidance such as point cloud diffusion \cite{chen2023text} or multi-view diffusion \cite{hong2024debiasing, shi2023mvdream, liu2023zero, armandpour2023reimagine} to maintain a global structural consistency.
However, they fail to straightway capture the high-order correlations within various aspects of 3D objects, such as the texture of symmetrical or correlated parts, ultimately compromising the fidelity and usability of the generated models.

To tackle this challenge, our research proposes a designed framework, ``3D Gaussian Generation via Hypergraph (Hyper-3DG)'', effectively engineered to address the complex, high-order correlations that underpin the geometry and texture of 3D objects.
Our approach encompasses a primary workflow (``Mainflow'') alongside a critical module, the ``Geometry and Texture Hypergraph Refiner (HGRefiner)''.
At its core, our methodology adheres to a well-established pipeline, leveraging the capabilities of the pre-trained 3D generators and 2D diffusion models to initial and optimize the representations of 3D objects.
Specifically, guided by the pre-trained 2D diffusion models, Denoising Diffusion Implicit Model (DDIM) \cite{ho2020denoising, song2020score}, our process begins with a ``Warm-Up'' phase,  where we harness the power of a pre-trained 3D generator (\eg, Point-E \cite{nichol2022point}, Shap-E \cite{jun2023shap}) to generate a preliminary 3D object from textual descriptions. 
Upon obtaining this initial 3D object, our uniquely designed ``HGRefiner'' embarks on processing the 3D Gaussians by patchifying \cite{peebles2023scalable} them into smaller, more manageable patch-level 3D Gaussian clusters, subsequently rendering them into 2D images.
We then apply a pre-trained 2D image feature extractor (\eg, ResNet \cite{he2016deep}, ResNeXt \cite{xie2017aggregated}, ViT \cite{dosovitskiy2020image}, Swin-T \cite{liu2021swin}, Dino \cite{caron2021emerging}) to capture the latent visual features of these 2D images, thereby enriching the 3D object's representation.
This process not only improves patch-level semantic visual comprehension but also retains the rendering speed advantage inherent in 3D Gaussian Splatting.
Through this innovative process, our HGRefiner adeptly establishes high-order correlations within the physical spatial space as well as the latent visual space of the 3D objects at the patch level, facilitated by hypergraph learning \cite{feng2019hypergraph, gao2022hgnn+}, named as ``High-Order Refine'' phase.
In this way, the HGRefiner module refines the spatial and latent representations of 3D objects in a high-order, correlative manner, focusing on each individual part of the 3D objects.
Furthermore, our methodology ensures consistency in the initialization process and hypergraph refinement by employing the same evaluation metric, the Interval Score Matching (ISM) loss \cite{liang2023luciddreamer}.
By maintaining the same loss, we effectively prevent the deterioration of these fundamental characteristics throughout the refinement process, ensuring the integrity and fidelity of the 3D objects remain intact.
This meticulous refinement culminates in the generation of finely detailed 3D objects in response to textual prompts.

The key contributions of our work are summarized as follows:
\begin{enumerate}
    \item We propose a designed framework to address high-order correlations within 3D objects, aiming to optimize both the geometry and texture of the generated 3D objects. To our knowledge, this represents the first trial of its kind in tackling these intricate correlations in the task of 3D generation;
    \item The high-order correlative optimizing approach refines 3D Gaussians by fine-tuning both explicit attributes and latent visual features at a manageable, patch-level scale.
    \item Our proposed method is designed for low coupling and is capable of significantly improving the performance of 3D generation without adding to the computational load for the various backbone models.
\end{enumerate}

\section{Related Work}
\label{sec:rela}

In this section, we provide a concise overview of recent progress in the field of text-to-3D generation, 3D representations, and hypergraph learning.

\subsection{Text-to-3D Generation}

Early attempts at text-to-3D generation primarily utilized CLIP \cite{radford2021learning} as a guidance mechanism for optimization, often producing suboptimal results.
To harness the powerful generative capabilities of diffusion models, Zero-1-to-3 \cite{liu2023zero} fine-tuned a pre-trained 2D diffusion model conditioned on camera parameters to elicit 3D priors from the 2D diffusion model.
3D assets were then reconstructed from the generated multi-view images.
MVDream \cite{shi2023mvdream} proposed a multi-view diffusion framework to generate consistent multi-view images for 3D object synthesis.
Wonder3D \cite{long2023wonder3d} adapted a pre-trained 2D diffusion model into a cross-domain diffusion model to produce paired RGB images and normal maps, subsequently fusing them into textured meshes.
In addition to fine-tuning diffusion models on 3D datasets to generate explicit 3D guidance, another line of research has focused on utilizing a pre-trained 2D diffusion model to directly optimize 3D representations. These approaches usually incorporate differentiable 3D representations such as NeRF \cite{mildenhall2020nerf}, NeuS \cite{wang2021neus}, \etc, and optimize their parameters through backpropagation. 
DreamFusion \cite{poole2022dreamfusion} proposed Score Distillation Sampling (SDS) to sample 3D parameters by optimizing a distillation loss.
Score Jacobian Chaining \cite{wang2023score} offered an alternative formulation and arrived at similar parametrizations as SDS.
ProlificDreamer \cite{wang2024prolificdreamer} analyzed the objective function of SDS and proposed a particle-based variational framework named Variational Score Distillation (VSD) that significantly improved the quality of generated content.
Recent works have incorporated SDS with Gaussian Splatting \cite{kerbl20233d} to achieve faster optimization.
DreamGaussian \cite{tang2023dreamgaussian} proposed a multi-stage framework that optimizes coarse 3D Gaussians via SDS in the first stage, with meshes and UV maps extracted and refined subsequently.
GSGEN \cite{chen2023text} utilized the explicit representation of 3D Gaussians and applied a point cloud diffusion model for global geometric guidance. 
GaussianDreamer \cite{yi2023gaussiandreamer} focused on the initialization stage and proposed an augmentation strategy to improve performance.
LucidDreamer \cite{liang2023luciddreamer} analyzed the SDS loss and proposed Interval Score Matching (ISM) to tackle the over-smoothness and inconsistency issues of the original SDS method.
In this work, we empirically follow the well-established mainstream architecture approachs \cite{liang2023luciddreamer, wang2024prolificdreamer, chen2023text} for 3D Gaussian generation as the primary workflow of our method.
Building upon this mainstream pipeline, we further optimize the refiner component by employing a specially designed patch-level 3D Gaussian hypergraph neural network.



\subsection{3D Representations}

Differentiable 3D representations play a pivotal role in optimization-based 3D generation.
One of the most commonly used representations is Neural Radiance Fields (NeRF) \cite{mildenhall2020nerf}, which represents a 3D scene as a continuous function mapping 5D coordinates (3D spatial coordinates and 2D viewing direction) to volume density and view-dependent emitted radiance.
NeuS \cite{wang2021neus} adopts a set of signed distance functions (SDFs) to represent the surface of 3D objects.
Plenoxels (plenoptic voxels) \cite{Fridovich2022Plenoxels} represent 3D scenes via a 3D grid of spherical harmonics, enabling faster optimization than NeRF.
Recently, 3D Gaussian Splatting \cite{kerbl20233d} has emerged as a promising approach for balancing optimization speed, rendering quality, and rendering speed. 
This method utilizes clusters of 3D Gaussians to explicitly represent a 3D scene, offering a wide range of flexible options for scene manipulation and rendering.
In our work, we focus on generating high-quality 3D Gaussians, aligning with the current popular choices and recent mainstream approaches.

\subsection{Hypergraph Learning}

Hypergraph Learning \cite{gao2020hypergraph, gao2022hgnn+, bai2021hypergraph}has emerged as an effective approach for modeling complex relational data.
Traditional graph learning methods \cite{kipf2016semi, velivckovic2017graph} are limited to pairwise relationships, while hypergraphs provide a natural way to represent higher-order interactions among multiple entities.
Hypergraph neural networks (HGNNs) \cite{feng2019hypergraph} generalize graph convolution operations to hypergraph structures, allowing for the propagation of information along hyperedges.
Several works have explored HGNNs for tasks such as node classification \cite{yu2012adaptive,ma2021hyperspectral}, regression \cite{di2021hypergraph, di2022generating}, link prediction \cite{yadati2020nhp,li2013link,fan2021heterogeneous}, matching \cite{liao2021hypergraph}, 3D retrieval \cite{gao2011camera,feng2023hypergraph}, and clustering \cite{purkait2016clustering,li2017inhomogeneous}.
For 3D data, hypergraph learning offers a promising direction for modeling the higher-order relationships inherent \cite{gao20123, zhang2020hypergraph, nong2022adaptive, jiang2022hypergraph}.
By representing objects and their relationships as hypergraphs, it is capable of capturing complex interactions and generate coherent 3D scenes.
In this work, we further investigate the integration of hypergraph representations with 3D generation techniques, which remains an unexplored area of research.

\section{Method}
\label{sec:method}

\begin{figure}
    \centering
    \includegraphics[width=\linewidth]{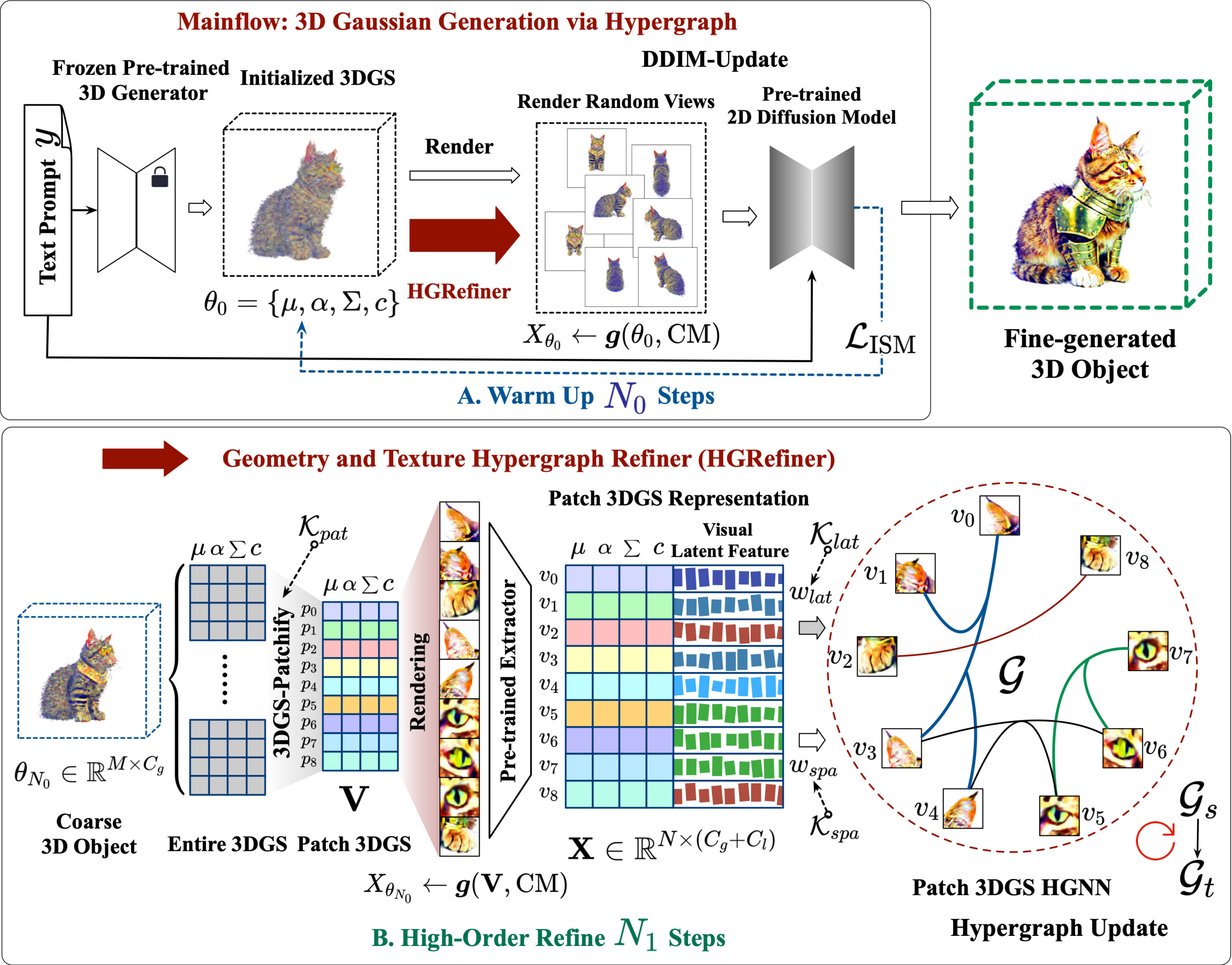}
    \caption{
    Illustration of the proposed method, 3D Gaussian Generation via Hypergraph (Hyper-3DG).
    Our method comprises a main flow as well as a designed hypergraph refiner module (Geometry and Texture Hypergraph Refiner).
    Given the text prompt as input, the ``Warm up'' stage can yield coarse 3D Gaussian by a pre-trained 3D generator and a 2D diffusion model.
    After $N_0$ steps of initialization, the ``HGRefiner'' further refines the geometry and texture of the coarse 3D Gaussian at the patch level, with an adjustable updated hypergraph structure.
    Following $N_1$ steps of high-order refinement, the final fine-generated 3D Object is obtained.}
    \label{fig:framework}
\end{figure}

Our objective is the creation of 3D content that boasts both precise geometry and rich detailing.
To achieve this, our approach, 3D Gaussian Generation via Hypergraph (Hyper-3DG), as illustrated in Fig.~\ref{fig:framework}, leverages the versatility of 3D Gaussian \cite{kerbl20233d} as a representational form.
This allows for the integration of geometric priors and the depiction of intricate high-frequency details.
Our method consists of two primary stages, namely ``Mainflow: 3D Gaussian Generation via Hypergraph'' and ``Geometry and Texture Hypergraph Refiner (HGRefiner)''.
Specifically, the pseudo codes are depicted in Algorithm~\ref{alg:main} and Algorithm~\ref{alg:hgrefiner}, respectively.

\begin{algorithm}[tb]
\SetKwInOut{To}{to}
\SetKwInOut{Input}{Input}
\SetKwInOut{Output}{Output}
\SetKwInOut{Initialization}{Initialization}
\SetKwFunction{FMain}{DDIM-Update}
\SetKwProg{Fn}{Function}{:}{}
\DontPrintSemicolon
\caption{Mainflow: 3D Gaussian Generation via Hypergraph}
\label{alg:main} 
\Input{
$y, N_0, N_1$,
$\mathrm{CM}$
}
\Output{
$\boldsymbol{\theta}$
}
\Initialization{The adjacent time step interval $\Delta t$ and 
the step size $\Delta s$ for predicting the noise trajectory in DDIM inversion.}
\BlankLine
\BlankLine
$\theta_0 \gets \textrm{Frozen-Pre-trained-3D-Generator}\left ( y \right )$
\tcp*[e]{Initialized Coarse 3DGS}
\For{$\theta_i \gets \theta_0$ \KwTo $i = N_0$}{
    \tcp{Update the 3DGS $\theta_i$ via pre-trained 2D diffusion model}
    $\theta_{i+1} \gets \textrm{DDIM-Update}(y, \theta_i, \textrm{CM})$
}
$\theta_i \gets \theta_{N_0}$
\tcp*[e]{Feed the obtained coarse 3DGS $\theta_{N_0}$ to HGRefiner}
\While{$\theta_i$ is not converged}{
    $\theta_j \gets \theta_i$\;
    \tcp{Keep the same Hypergraph sturcture for $N_1$ steps}
    \For{$j \gets 0$ \KwTo $j = N_1$}{
        $\widetilde{\theta}_j \gets \textrm{HGRefiner}(\theta_j, j, N_1)$
        \tcp*[e]{Update $\theta_j$ by HGRefiner}
        $\widehat{\theta_{j}} \gets \textrm{DDIM-Update}(y, \widetilde{\theta}_j, \textrm{CM})$
    }
    $\theta_i \gets \widehat{\theta_{j}}$
}

$\boldsymbol{\theta} \gets \theta_i$
\tcp*[e]{Obtain the final Fine-generated 3D Object}

\BlankLine  

\Fn{\FMain{$y, \theta_i, \mathrm{CM}$}}{
    $X_{\theta_i} \gets \boldsymbol{g} \left (\theta_i, \mathrm{CM} \right )$
    \tcp*[e]{Render 2D images from the 3DGS}
    
    $t \sim \mathcal{U}(1, \lambda), \lambda \in \mathbf{R}^+, s = t - \Delta t$
    \tcp*[e]{Timestep $t$ and the pre-timestep $s$}
    
    $X_{\theta_i}^t, X_{\theta_i}^s \gets \textrm{DDIM}(y, X_{\theta_i}, t, s, \Delta s)$ \;
    \tcp{$X_{\theta_i}^t, X_{\theta_i}^s$ denote the noisy latent vector at $t, s$, respectively}
    
    $\nabla_\theta\mathcal{L}_\mathrm{ISM}(\theta_i) \gets \mathcal{L}_\mathrm{ISM}({\epsilon}_\phi(X_{\theta_i}^t,t,y),{\epsilon}_\phi(X_{\theta_i}^s,s,\emptyset))$
    \tcp*[e]{Gradient Descent}
    
    \If{$\theta_i$ is not converged}{
        $\widehat{\theta_{i}} \gets \nabla_\theta\mathcal{L}_\mathrm{ISM}(\theta_i)$
        \tcp*[e]{Update $\theta_i$  with loss function $\mathcal{L}_{\mathrm{ISM}}$}
    }
    \KwRet $\widehat{\theta_{i}}$
}
\end{algorithm}

\subsection{Mainflow: 3D Gaussian Generation via Hypergraph}

In this section, we elaborate on the main-flow of our method.
In the initial phase, named as ``Warm-up'' process in Fig.~\ref{fig:framework}, the objective is to employ a frozen pre-trained 3D generative model and a 2D pre-trained Diffusion model to establish the preliminary geometry and texture of the 3D object from the specified text prompt.
The initial establishment of the 3D objects, as described, lays the groundwork for subsequent refinement and enhancement of details.
This trunk process is widely embraced as an empirical practice within the field of text-to-3D object generation (\eg, \cite{liang2023luciddreamer, wang2024prolificdreamer, ma2023geodream, sun2023dreamcraft3d}).

Beginning with a textual prompt $y$, we first obtain a rough version of the 3D Gaussian from scratch using a frozen, pre-trained point cloud Generator (\eg, Point-E \cite{nichol2022point}, Shap-E \cite{jun2023shap}).
We denote this initialized 3D Gaussian Splatting (3DGS) as $\theta_0 = \{ \mu, \alpha, \Sigma, c \}$, where $\mu, \alpha, \Sigma, c$ respectively represent the mean (\ie, center position (\textit{x, y, z})), opacity, covariance, and view-dependent color of each corresponding 3D Gaussian distribution.

Subsequently, we introduce a module named ``DDIM-Update'', which employs a pre-trained 2D Diffusion Model (\eg, Denoising Diffusion Implicit Model \cite{ho2020denoising, song2020score}) to optimize and refine the 3D Gaussian distribution.
This development is inspired by and derived from the methodologies outlined in LucidDreamer \cite{liang2023luciddreamer}.
The module DDIM-Update takes the text prompt ($y$), 3D Gaussian distributions ($\theta_0$) and the camera poses ($\textrm{CM}$) as inputs to deliver a more reliable and consistent trajectory for the latent state.
The initial 2D images $X_{\theta_0} = \{ \boldsymbol{x}_0, \boldsymbol{x}_1, \cdots\} \in \mathbb{R}^{\| \textrm{CM} \|}$ are rendered by $X_0 = \boldsymbol{g}(\theta_0, \textrm{CM})$ with the rendering function $\boldsymbol{g}(\cdot)$ as well as the random camera poses $\textrm{CM} = \{cm_0, cm_1, \dots\}$.
Then, the DDIM \cite{song2020denoising} inversion transforms the 2D images into a sequence of unconditional noisy latent trajectories $\{X_{\theta_i}^{\Delta t}$, $X_{\theta_i}^{2\Delta t}$, $\dots$, 
$X_{\theta_i}^s$, $X_{\theta_i}^t \},~X_{\theta_i}^s = X_{\theta_i}^{t - \Delta t}$ is the noise sequence at step $s$ derived from the input $X_{\theta_i}$, where $\Delta t$ is the DDIM inversion step size and the notation $X_{\theta_i}$ for $i \in [0,N_0-1]$ represents the $i-th$ update during the warmup phase across $N_0$ steps, including diverse views rendered from different camera pose $cm_j$ within the range $j \in [0,\| \textrm{CM} \|-1]$.
Considering $\epsilon_\phi(\cdot)$ as the predicted noise by the 2D diffusion, the iterative prediction can be formulated as:
\begin{equation}\boldsymbol{x}_t=\sqrt{\bar{\alpha}_t}\bar{\boldsymbol{x}}_0^s+\sqrt{1-\bar{\alpha}_t}{\epsilon}_\phi(\boldsymbol{x}_s,s,y); ~\boldsymbol{x}_t \in X_{\theta_i}^t,~\boldsymbol{x}_s \in X_{\theta_i}^s;~y = \emptyset \end{equation}
where $s = t - \Delta t$, $\bar{\boldsymbol{x}}_0^s=\bar{\alpha}_s^{-\frac{1}{2}}\boldsymbol{x}_s-\gamma(s){\epsilon}_\phi(\boldsymbol{x}_s,s,y=\emptyset)$, the condition $y$ here is emptyset $\emptyset$, and $\bar{\alpha}_{t}$ is a function of the diffusion schedule.
During the update phase, we leverage the Interval Score Matching (ISM) loss $\mathcal{L}_{\mathrm{ISM}}$ \cite{liang2023luciddreamer} to reduce the difference between the denoising directions at two distinct intervals within the diffusion trajectory, which is mathematically expressed as:
\begin{align}
\mathcal{L}&\triangleq\mathbb{E}_{t,c}\left[\omega(t)||{\epsilon}_{\phi}(\boldsymbol{x}_{t},t,y)-{\epsilon}_{\phi}(\boldsymbol{x}_{s},s,\emptyset)||^{2}\right] \label{eq:ism-loss} \\
\nabla_\theta\mathcal{L}_\mathrm{ISM}(\theta) &= \mathbb{E}_{t,c}\left[\omega(t)(\underbrace{{\epsilon}_\phi(\boldsymbol{x}_t,t,y)-{\epsilon}_\phi(\boldsymbol{x}_s,s,\emptyset)}_\mathrm{ISM~opt~direction})\frac{\partial\boldsymbol{g}(\theta,c)}{\partial\theta}\right] \label{eq:ism-gradient}
\end{align}

Specifically in practice, we adopt the approach of enhancing efficiency by forecasting $X_{\theta_i}^s$ with large step size $\Delta s$ in the multi-step DDIM denoising process.
The ISM loss guides the optimization of the 3D model's parameters $\theta$ to produce detailed and realistic 3D objects, effectively overcoming the over-smoothing problem associated with traditional SDS methods.

Upon completion of $N_0$ iterations, the ``Warm Up'' phase yields the parameter set $\theta_{N_0}$, which is subsequently passed to the HGRefiner stage for additional optimization.

\subsection{Geometry and Texture Hypergraph Refiner}

\begin{algorithm}[tb]
\SetKwInOut{Input}{Input}
\SetKwInOut{Output}{Output}
\DontPrintSemicolon
\caption{Geometry and Texture Hypergraph Refiner (HGRefiner)}
\label{alg:hgrefiner}
\Input{
$
\theta_{j} \in \mathbb{R}^{M \times C_g},
\mathcal{K}_{pat},
\mathcal{K}_{spa},
\mathcal{K}_{lat},
j,
N_1
$
}
\BlankLine
\Output{
$\widetilde{\theta}_{j} \in \mathbb{R}^{M \times C_g}$
}
\BlankLine
\BlankLine

$\mathbf{V}_j \gets \textrm{3DGS-Patchify}\left (\theta_j, \mathcal{K}_{pat}\right )$
\tcp*[e]{Yield the patch-level 3DGS}
$X_j \gets \boldsymbol{g} \left (\mathbf{V}_j, \mathrm{CM} \right )$
\tcp*[e]{Render the 2D images from patch-level 3DGS}
$\mathbf{F}_j \gets \textrm{2D-Img-Extractor} \left ( X_j \right )$
\tcp*[e]{Extract the latent visual feature}
$\overline{\mathbf{V}}_j \gets \textrm{Mean-in-Patch} \left( \mathbf{V} \right)$
\tcp*[e]{Mean vector of each patch-level 3DGS}
$\mathbf{X}_j \gets \overline{\mathbf{V}}_j || \mathbf{F}_j $
\tcp*[e]{Concatenate explicit and latent representation}
\If{$j == N_1$}{
    \tcp{Update the structure of Patch-3DGS Hypergraph}
    $\mathcal{G}^{spa}_j = \langle \mathbf{X}_j, \mathbf{H}_j^{spa}, \mathbf{W} \rangle \gets \textrm{Construct-Patch-3DGS-Hypergraph} \left ( \mathbf{X}_j, \mathcal{K}_{spa} \right )$\;
    $\mathcal{G}^{lat}_j = \langle \mathbf{X}_j, \mathbf{H}_j^{lat}, \mathbf{W} \rangle \gets \textrm{Construct-Patch-3DGS-Hypergraph} \left ( \mathbf{X}_j, \mathcal{K}_{lat} \right )$\;
}
\Else{
\tcp{Not update the structure of Patch-3DGS Hypergraph}
$\mathcal{G}^{spa}_j \gets \mathcal{G}^{spa}_{j-1}$
\tcp*[e]{Get the previous hypergraph from last step}
$\mathcal{G}^{lat}_j \gets \mathcal{G}^{lat}_{j-1}$
\tcp*[e]{Get the previous hypergraph from last step}
}
$\widetilde{\mathbf{X}}_j \gets \textrm{Patch-3DGS-HGNN} \left( \mathbf{X}_j, \mathcal{G}^{spa}_j, \mathcal{G}^{lat}_j, w_{spa}, w_{lat} \right )$ \;
$\Delta \theta_j \gets \textrm{3DGS-Recover} \left ( \widetilde{\mathbf{X}}_j - \mathbf{X}_j \right )$
\tcp*[e]{Calculate updates, recover shape}
$\widetilde{\theta}_j \gets \theta_j + \Delta \theta_j $
\tcp*[e]{Obtain the Optimized entire 3DGS}

\end{algorithm}

During this stage, the ``HGRefiner'' module takes as its input the coarse 3D Gaussian distribution generated in the previous ``Warm Up'' stage and improves its quality, specifically the geometry and texture, through the designed ``Patch 3DGS Hypergraph Learning''.
We represent the entire 3D Gaussian of the coarse 3D object as $\theta_{N_0} \in \mathbb{R}^{M \times C_g}$, where $M$ and $C_g$ represent the number of Gaussian distributions and the attributes of the 3DGS ($\mu, \alpha, \Sigma, c$), respectively.
The employment of 3D Gaussian distribution as a method for 3D object representation involves handling extensive data volumes, which complicates the extraction of latent semantic visual representations.
To address this challenge, we introduce a mechanism termed ``3DGS-Patchify'', designed to compress and reduce the 3D Gaussian to patch-level affordable dimensions, in spatial space.

In this context, we employ the K-Means clustering algorithm \cite{krishna1999genetic} as the implementation mechanism for 3DGS-Patchify.
This approach segments the entire 3DGS ($\mathbb{R}^{M \times C_g}$) into $N$ clusters, represented as $\mathbf{V} \in \mathbb{R}^{\left(N \cdot \frac{M}{N} \right) \times C_g}$, where $N \ll M$ signifies the number of clustered patch-level 3DGS.
Note that the scale of patchify number $N$ is governed by a hyper-parameter $\mathcal{K}_{pat}$ (where $N \Leftarrow \mathcal{K}_{pat}$ in K-Means).
Each patch-level 3DGS represents a small cluster of 3D Gaussian and can be rendered into a patch-level 2D image by the render function ($\boldsymbol{g}(\cdot)$) using specified camera parameters $\textrm{CM}$.
We denote these patch-level 2D images as $X_{\theta_{N_0}} = \{\boldsymbol{x}_1, \boldsymbol{x}_2, \cdots, \boldsymbol{x}_N \} \in \mathbb{R}^{N}$.

Upon acquiring the patch-level 2D images ($X_{\theta_{N_0}}$) and their corresponding patch-level 3DGS, we subsequently treat them as vertices and construct the Patch 3DGS Hypergraph.
The tensor representation of these vertices ($\mathbf{X} = \overline{\mathbf{V}} || \mathbf{F} $) is obtained by concatenating the mean vector of each explicit attribute of the patch-level 3DGS ($\overline{\mathbf{V}} \in \mathbb{R}^{N \times C_g}$) with the latent visual features ($\mathbf{F}\in \mathbb{R}^{N \times C_l}$) derived from a pre-trained 2D image feature extractor (\eg, ResNet \cite{he2016deep}, ResNeXt \cite{xie2017aggregated}, ViT \cite{dosovitskiy2020image}, Swin-T \cite{liu2021swin}).
We denote this tensor representation as $\mathbf{X} \in \mathbb{R}^{N \times (C_g + C_l)}$, where $C_g$ and $C_l$ denote the dimension of 3DGS attributes and the latent visual features, respectively.
Considering the representation contains the explicit spatial information and the latent semantic features, we consequently construct the spatial hypergraphs ($\mathcal{G}^{spa}$) as well as semantic hypergraphs ($\mathcal{G}^{lat}$) with corresponding dynamic weights $w_{spa}$ and $w_{lat}$.
The construction of the hypergraph can be implemented as the K-nearest neighbors (KNN) algorithm \cite{peterson2009k}, applied separately ($\mathcal{K}_{spa}, \mathcal{K}_{lat}$) in both spatial ($\mu \Leftrightarrow (x, y, z)$) and latent spaces ($\mathcal{F}$), through calculating the Euclidean distance.
In this way, the spatial and latent Patch 3DGS hypergraphs are constructed and respectively denoted as $\mathcal{G}^{spa} = \langle \mathbf{X}, \mathbf{H}^{spa}, \mathbf{W} \rangle$ and $\mathcal{G}^{lat} = \langle \mathbf{X}, \mathbf{H}^{lat}, \mathbf{W} \rangle$.
We denote $\mathbf{H}^{(\cdot)} \in \mathbb{R}^{N \times E}$ and $\mathbf{W} = \mathbf{1} \in \mathbb{R}^{E \times E}$ respectively as the incidence matrix and the vertices weight matrix (\eg, all-ones matrix $\mathbf{1}$), where $E$ represents the number of hyperedges in the hypergraph.
The ``Patch-3DGS-HGNN'' referred to the line 12 of Algorithm~\ref{alg:hgrefiner} is formulated as follows:
\begin{equation}
\left\{
\begin{matrix}
\begin{aligned} 
& \mathbf{H} = \mathbf{H}^{spa} \| \mathbf{H}^{lat}  \\
& \widetilde{\mathbf{X}}=\sigma\left(\mathbf{D}_v^{-1 / 2} \mathbf{H} \mathbf{W} \mathbf{D}_e^{-1} \mathbf{H}^{\top} \mathbf{D}_v^{-1 / 2} \mathbf{X} \boldsymbol{\mathbf{\Theta}}\right)
\end{aligned}
\end{matrix}
\right.
\end{equation}
where $\mathbf{D}_e \in \mathbb{R}^{E \times E}$, $\mathbf{D}_v \in \mathbb{R}^{N \times N}$ and $\mathbf{W} \in \mathbb{R}^{E \times E}$ denote the diagonal degree matrix of hyperedges, the degree matrix of vertices and weight matrix of hyperedges, respectively.
$\sigma(\cdot)$ denotes the nonlinear activation function (\eg, $\mathrm{LeakyReLU}(\cdot)$).
$\mathbf{\Theta} \in \mathbb{R}^{(C_g + C_l) \times (C_g + C_l)}$ is a diagonal matrix representing the learnable parameters updated by the ISM loss function in the outer loop (\ie, Mainflow).
It functions similarly to a multilayer perceptron (MLP) layer.
By calculating the difference between the original representation tensor ($\mathbf{X} \in \mathbb{R}^{N\times (C_g + C_l)}$) and the updated representation tensor ($\widetilde{\mathbf{X}} \in \mathbb{R}^{N\times (C_g + C_l)}$), and by employing the ``3DGS-Recover'' function, we generate the update amounts for patch-level 3DGS ($\Delta \theta \in \mathbb{R}^{M \times C_g}$).
The function ``3DGS-Recover'' simply drops the latent visual features ($\mathbf{F}$) from the representation tensor ($\mathbf{X}, \widetilde{\mathbf{X}}$) and recovers the patch-level 3DGS to the original shape by replicating augmentation ($\left ( \bigotimes \mathbb{R}^{N} \to \mathbb{R}^M \right)$).
The final updated 3DGS is produced by adding the updating increments ($\Delta \theta \in \mathbb{R}^{M\times C_g}$) to the original 3DGS ($\theta \in \mathbb{R}^{M\times C_g}$).

Upon completing $N_1$ steps of this High-Order Patch-3DGS refinement, coupled with the main flow, and ensuring the entire 3DGS ($\boldsymbol{\theta}$) has converged, the final finely detailed 3D object is generated.

\section{Experiments}
\label{sec:exp}

In this section, we elaborate our experiments conducted to validate the effectiveness of the proposed 3DGHG approach.
Specifically, we benchmark 3DGHG against previous state-of-the-art methods in the domain of text-to-3D generation.
Moreover, we conduct a series of ablation studies to evaluate the significance of crucial components within our method, encompassing each hyper-parameters, loss functions, pre-trained 2D and 3D models, and other relevant factors.
The comprehensive results of these investigations are presented hereafter.

\subsection{Comparison Experiment}

\subsubsection{Comparative Methods and Settings}

Comparative experiments are conducted in the domain of text-to-3D generation, with 3D Gaussian Splatting \cite{kerbl20233d} serving as the chosen representation for 3D objects.
To ensure a fair assessment, we employ identical textual prompts and consistent settings across all methods.
For instance, we utilize pre-trained 3D models (\ie, Point-E \cite{nichol2022point}) and a pre-trained 2D model (\ie, Stable Diffusion 2.1 \cite{rombach2022high}).
The Classifier-Free Guidance (CFG) parameter \cite{ho2022classifier} is set to 100 for the methods based on SDS \cite{poole2022dreamfusion, tang2023dreamgaussian, chen2023text}, and to 7.5 for the method employing the ISM loss \cite{liang2023luciddreamer}.
All models are trained for 4,000 epochs.
Other parameters are adjusted in line with the respective official methodologies \cite{poole2022dreamfusion, tang2023dreamgaussian, chen2023text, liang2023luciddreamer} to ensure apples-to-apples comparisons.

In our comparative analysis, we benchmark our method against the following state-of-the-art approaches:
\begin{itemize}
\item \textbf{DreamFusion} \cite{poole2022dreamfusion} is an optimization-based method that lifts 2D content to 3D. It introduces the Score Distillation Sampling (SDS) technique, which leverages pre-trained 2D diffusion models to generate 3D content.
\item \textbf{DreamGaussian} \cite{tang2023dreamgaussian} employs a three-stage optimization process from coarse to fine detail. Initially, SDS and 3D Gaussians are used to quickly generate a coarse 3D representation. This is followed by the extraction of meshes, which are then used for UV map refinement in the final stage.
\item \textbf{GSGEN} \cite{chen2023text} integrates SDS with 3D Gaussians. It takes advantage of the explicit nature of 3D Gaussians and applies a point cloud diffusion model as global geometric guidance to the generated 3D objects, addressing the multi-face Janus problem.
\item \textbf{LucidDreamer} \cite{liang2023luciddreamer} analyzes the SDS loss characteristics and introduces the Interval Score Matching (ISM) loss to counteract the excessive smoothness and lack of consistency in the original SDS loss. This innovation leads to a notable enhancement in the quality of the produced 3D Gaussians.
\end{itemize}

\begin{figure}[tb]
    \centering
    \includegraphics[width=\linewidth]{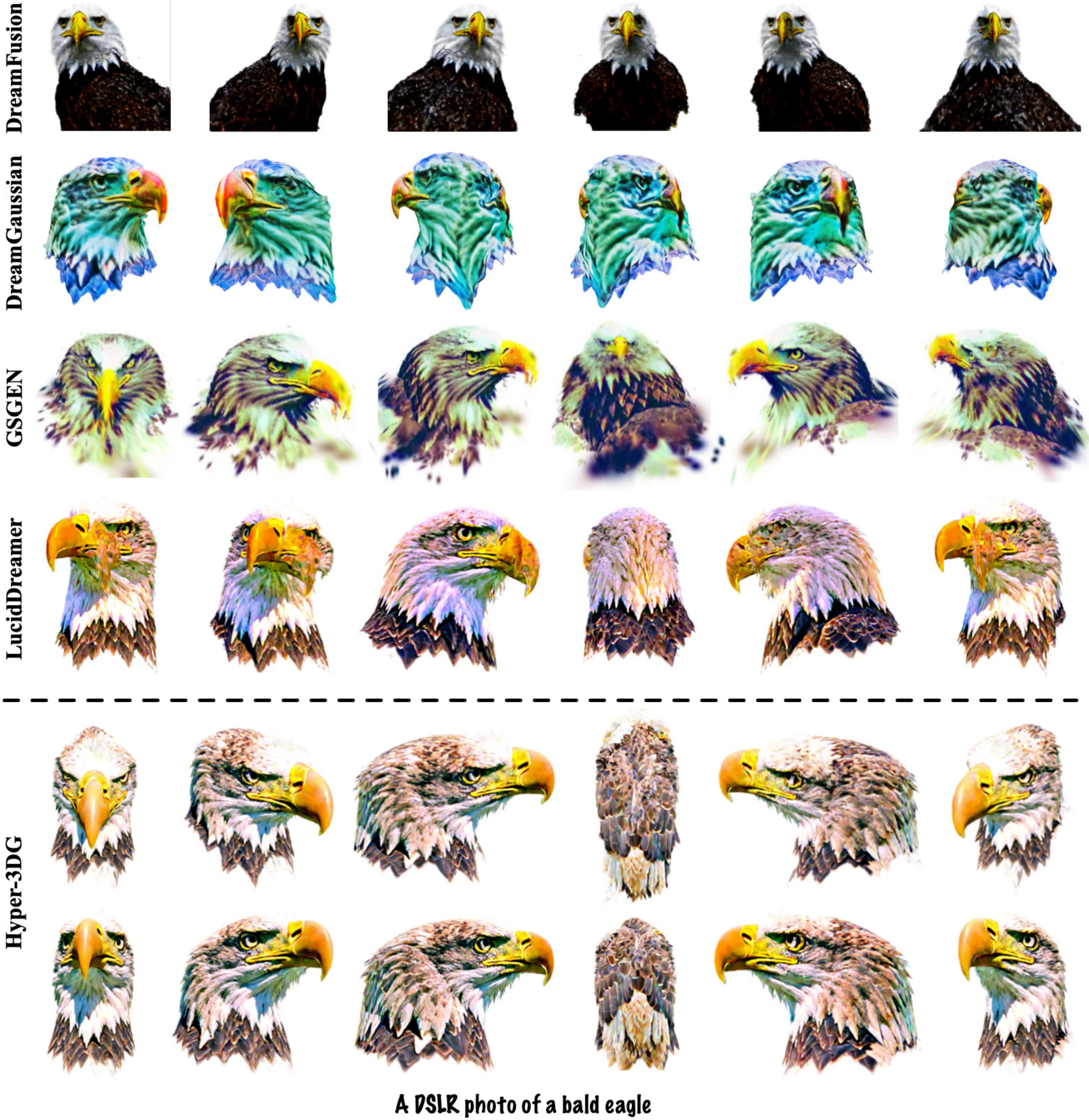}
    \caption{One example of DreamFusion \cite{poole2022dreamfusion}, DreamGaussian \cite{tang2023dreamgaussian}, GSGEN \cite{chen2023text}, LucidDreamer \cite{liang2023luciddreamer}, and our proposed method Hyper-3DG (the final two lines depicting contrasting perspectives, \ie, thophoric view and the overlook) with the same settings. The images in each column represent rendering results from an identical perspective. A few methods could not generate the back view known as the Janus problem. The results demonstrate the superiority of our approach in synthesizing highly realistic content, replete with intricate details. Please zoom in for the finer intricacies.}
    \label{fig:fig_exp_1}
\end{figure}

\begin{figure}[tb]
    \centering
    \includegraphics[width=\linewidth]{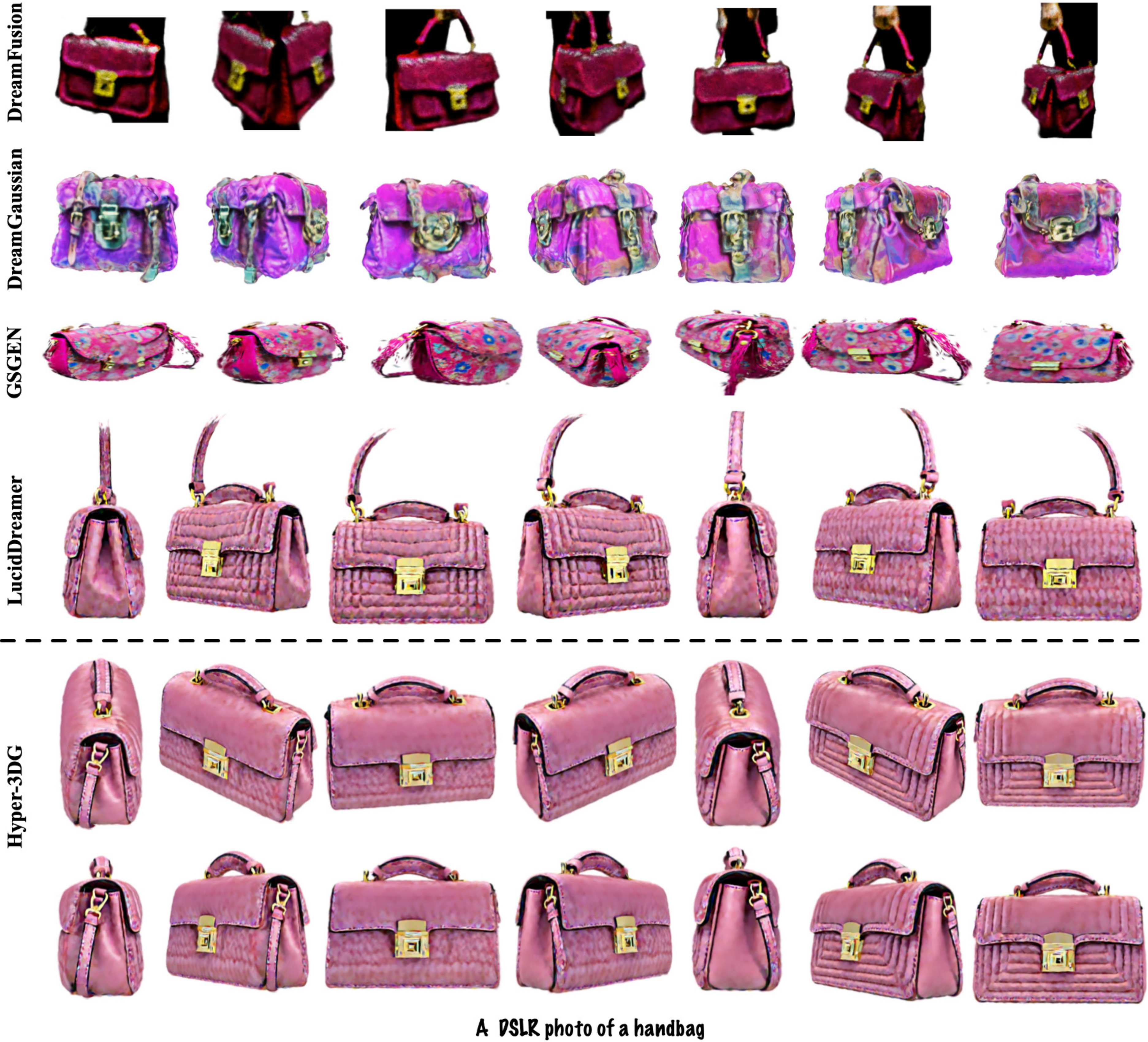}
    \caption{A comparison of experimental results among state-of-the-art methods and our approach under identical settings.}
    \label{fig:fig_exp_2}
\end{figure}

\begin{figure}[tb]
    \centering
    \includegraphics[width=\linewidth]{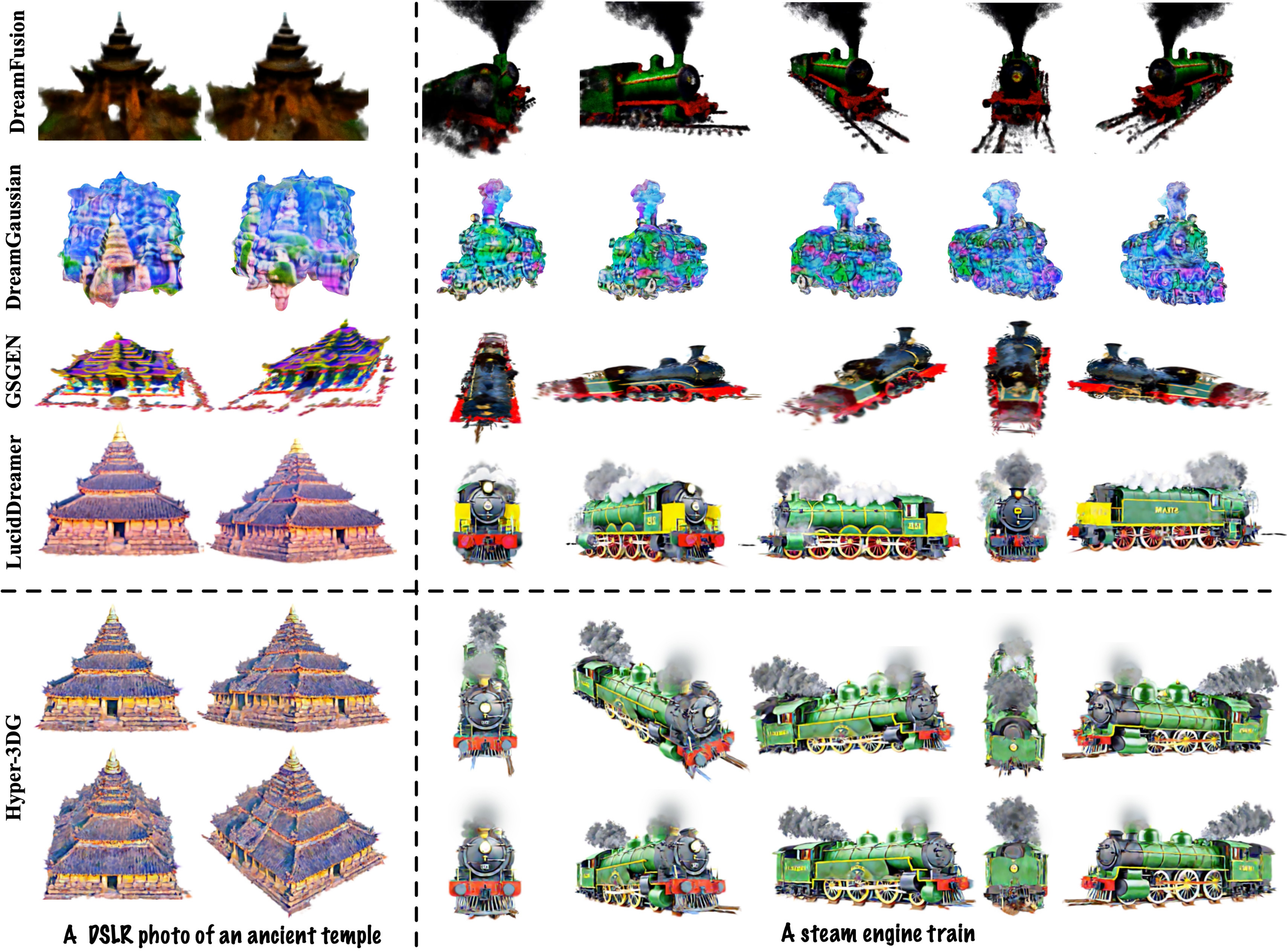}
    \caption{A comparison of experimental results among state-of-the-art methods and our approach under identical settings.}
    \label{fig:fig_exp_3}
\end{figure}

\subsubsection{Implementation Details}
\label{sec:imp-details}
All experiments are conducted using the stable diffusion model 2.1 for distillation purposes, and to ensure consistency and fairness in comparison, NVIDIA 4090 GPUs were used across all trials.
We utilize the official implementation of Point-E \cite{nichol2022point} to generate coarse 3D assets and integrate them into differentiable 3D representations.
After the initialization phase, we proceed to the ``Warm-Up'' phase and then apply our Hyper-3DG refinement (HGRefiner) stage.
The basic experimental setup is as follows:
\textit{iterations} = 4000,
$N_0$ = 1000,
$N_1$ = 50,
$\mathcal{K}_{pat}$ = 50,
$\mathcal{K}_{lat}$ = 13,
$\mathcal{K}_{spa}$ = 13,
and the 3DGS position learning rate is set to $1.6 \times 10^{-6}$, 2D image latent feature extractor is ViT.
During each $N_1$ iteration where our Hyper-3DG method is employed, the GPU memory consumption is approximately 3,570 MiB, and the process takes about 1.2 minutes to complete.
The processes of patchify and hypergraph construction are each completed in less than 10 seconds. The image rendering and feature extraction for the patchify 3DGS step typically require between 30 to 50 seconds. 
Subsequently, the final update of the hypergraph generally takes about 10 seconds to complete.




\subsubsection{Results and Analysis}
Following the implementation outlined in the previous sections, we present several comparative results, as depicted in Fig.~\ref{fig:fig_exp_1}, Fig.~\ref{fig:fig_exp_2}, and Fig.~\ref{fig:fig_exp_3}.
Based on these results, we can derive the following key observations:
\begin{itemize}
    \item \textbf{Enhanced cross-view consistency.} Our method achieves a higher level of view consistency in the generated objects, as demonstrated in the example of the eagle's beak shown in Fig.~\ref{fig:fig_exp_1}, where it outperforms other methods. The eagle beak produced by our method exhibits greater fidelity when viewed from different angles. Moreover, our method surpasses other approaches in preserving consistency between the front and back views, effectively addressing the Janus Problem. In contrast, the comparative methods, such as LucidDreamer \cite{liang2023luciddreamer}, may result in inconsistencies, with multiple beaks or missing eyes. Other methods like DreamFusion \cite{poole2022dreamfusion}, DreamGaussian \cite{tang2023dreamgaussian}, and GSGEN \cite{chen2023text} struggles to consistently generate the back view, yielding less satisfactory results;

    \item \textbf{Advanced color and texture.} Our method excels in generating 3D assets with highly natural and detailed color and texture. For instance, the feathers of the eagle in Fig.~\ref{fig:fig_exp_1} exhibit a more refined and lifelike appearance. The handbag in Fig.~\ref{fig:fig_exp_2} demonstrates a more authentic texture, particularly in the smooth and realistic depiction of the strap. The temple in Fig.~\ref{fig:fig_exp_3} showcases more precise structural and textural details. In the ``steam engine train'' example of Fig.~\ref{fig:fig_exp_3}, our method renders the wheels of train with greater roundness and uniformity, and the smoke appears more realistic. This stands in contrast to other methods \cite{poole2022dreamfusion, chen2023text, tang2023dreamgaussian}, which may suffer from over-smoothing or over-saturation, leading to the generation of unrealistic colors or the inability to produce valid colors, as seen in the comparison examples.

    \item \textbf{Improved structural integrity.} Our method successfully addresses the challenge of structural incoherence by effectively filling gaps between correlated structures. This is exemplified in Fig.~\ref{fig:fig_exp_2}, where our method produces a handbag with superior structural integrity. In comparison, the strap of the handbag generated by the LucidDreamer \cite{liang2023luciddreamer} method appears incomplete. This improvement is attributable to the capability of HGRefiner to refine and optimize geometric information, resulting in a complete and coherent structure. Other methods, as demonstrated, may fail to form a normal and complete geometry of the handbag, highlighting the superiority of our approach in maintaining structural integrity.

\end{itemize}

\subsection{Ablation Study}
This section employs the foundational parameters outlined in Section \ref{sec:imp-details} as the constants for the ablation study. Constrained by limited manpower and time, we only sample the intermediate rendering state display rather than the final results of the 4000 iterations.

\subsubsection{Ablation Study on Loss Function}

\begin{figure}[tb]
    \centering
    \includegraphics[width=\linewidth]{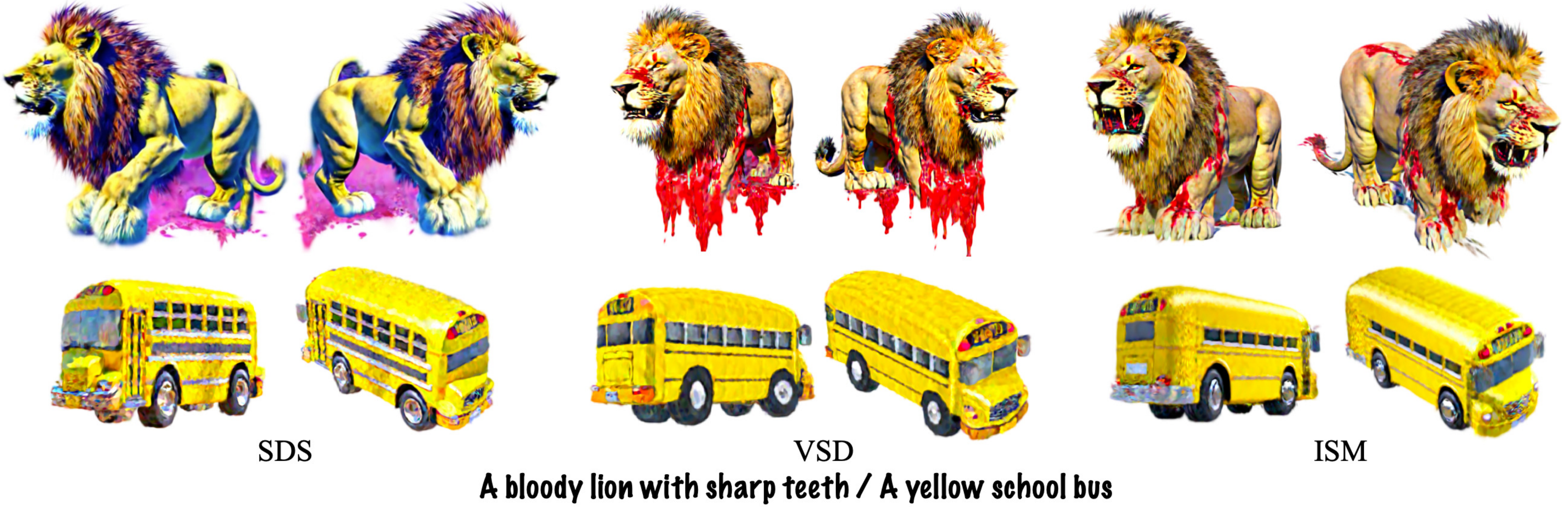}
    \caption{\textbf{Loss Function}. The comparative results of employing different loss functions (\ie, SDS \cite{chen2023text}, VSD \cite{wang2024prolificdreamer}, ISM \cite{liang2023luciddreamer}) within our proposed framework, with identical settings maintained across all experiments.}
    \label{fig:loss}
\end{figure}

To assess the effectiveness of various loss functions within our proposed framework, we conducted a comparative analysis with consistent settings across all experiments.
Several loss functions are commonly used for the task of text-to-3D generation.
Focusing on 3D Gaussian Splatting generation, we introduce and compare the experimental performance of three prominent loss functions:
SDS (Score Distillation Sampling) \cite{poole2022dreamfusion, chen2023text},
VSD (Variational Score Distillation) \cite{wang2024prolificdreamer},
and ISM (Interval Score Matching) \cite{liang2023luciddreamer}, which have been proposed and widely adopted in state-of-the-art methods.
\begin{itemize}
    \item \textbf{SDS}, introduced in DreamFusion \cite{poole2022dreamfusion}, also known as Score Jacobian Chaining \cite{wang2023score}, is a popular optimization-based sampling method for 3D asset generation.
    \item \textbf{VSD} \cite{wang2024prolificdreamer} builds upon SDS by incorporating a variational framework and simulating a Wasserstein gradient flow ODE to generate samples. It offers higher-quality samples but at a greater computational cost.
    \item \textbf{ISM} \cite{liang2023luciddreamer} improves upon SDS by replacing the noise term and noise prediction term with noise predictions from DDIM-inversed latents, providing better sample quality than SDS and lower computation cost than VSD.
\end{itemize}
The results presented in Fig.~\ref{fig:loss} indicate that ISM achieves superior texture quality and detail while maintaining computational efficiency.
For instance, the lion figure produced using SDS exhibits inaccurate color representation, with the red blood color appearing purple, whereas the colors of samples generated by ISM are more accurate. 
Additionally, the bus figure produced with ISM displays more detailed and realistic textures compared to those generated by VSD and SDS.
Based on this empirical evidence, we recommend adopting the ISM loss function in our framework, as it is likely to yield the best results with a high probability.

\subsubsection{Ablation Study on 3DGS-Patchify}

\begin{figure}[tb]
    \centering
    \includegraphics[width=\linewidth]{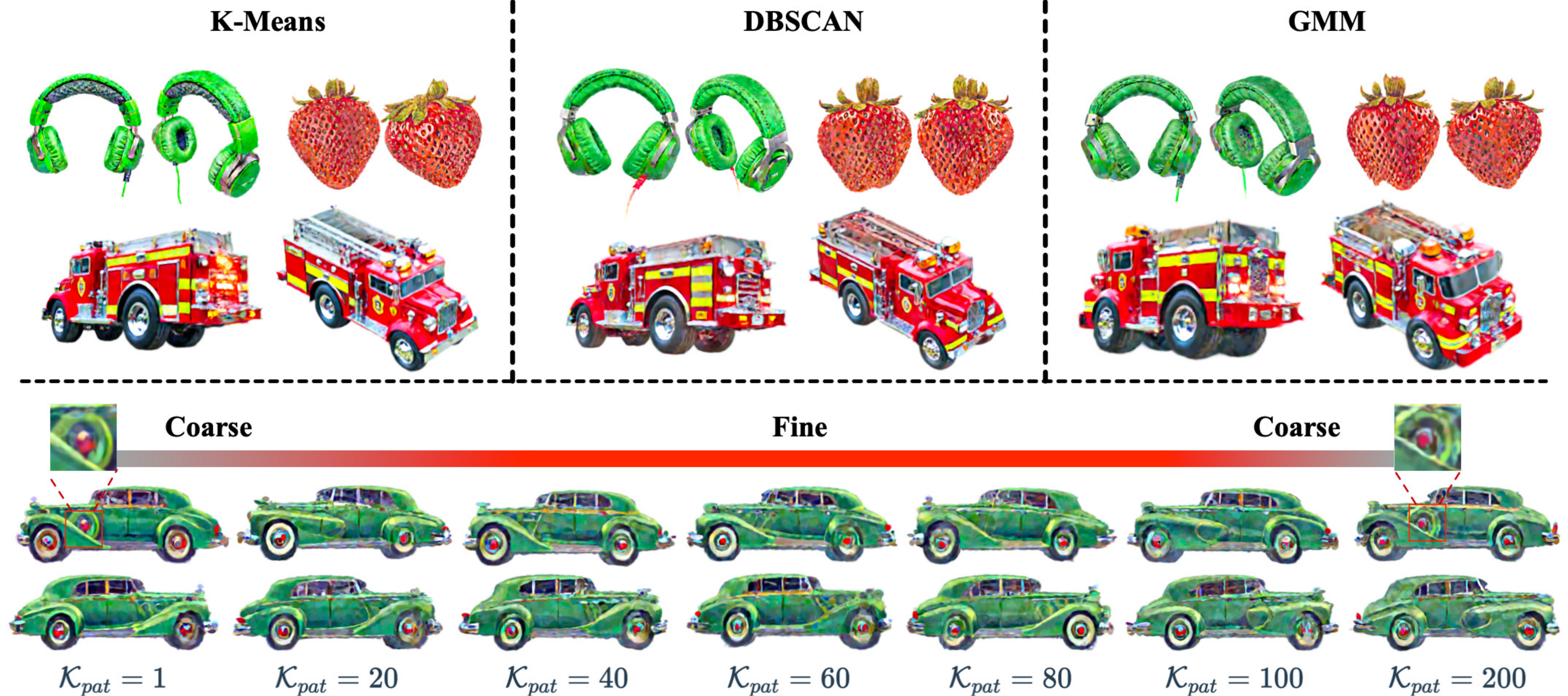}
    \caption{\textbf{3DGS-Patchify}. The comparative results of employing different $\textrm{3DGS-Patchify}$ functions (\ie, K-Means \cite{hamerly2003learning}, DBSCAN \cite{ester1996density}, GMM \cite{zhuang1996gaussian}) and the different hyper-parameter of K-Means (denoted as $\mathcal{K}_{pat}$) within our proposed framework, with identical settings maintained across all experiments. Here, the prompts are respectively ``A pair of green headphones'', ``A ripe strawberry'', ``A classic fire truck'', and ``A classic Packard car''.}
    \label{fig:patchify}
\end{figure}

To assess the effectiveness of various implementations of ``3DGS-Patchify'' and to determine the optimal hyper-parameter ($\mathcal{K}_{pat}$), we performed a comparative analysis under consistent experimental conditions.
As depicted in Fig.~\ref{fig:patchify}, both DBSCAN \cite{ester1996density} and GMM \cite{zhuang1996gaussian} yield less satisfactory results compared to K-Means \cite{hamerly2003learning} in the function of ``3DGS-Patchify''.
For instance, DBSCAN and GMM may produce incomplete strawberry shapes due to erroneous clustering outcomes.
Similarly, in the cases of the headphone and fire truck examples, these methods do not achieve the level of detail and realism offered by K-Means.
We further investigate the impact of the $\mathcal{K}_{pat}$ hyper-parameter of K-Means, by conducting an ablation experiment across a range of values from 1 to 200.
Our observations indicate that artifacts resembling tire shapes emerge on the body of car at the extreme values of $\mathcal{K}_{pat}$, specifically when $\mathcal{K}_{pat}$ is set to 1 or 200. 
Furthermore, a consistent pattern of artifacts appearing in the same locations is noticeable when $\mathcal{K}_{pat}$ is set to 80.
These findings suggest that the optimal range for the $\mathcal{K}_{pat}$ parameter may lie within the vicinity of 80, as both lower and higher values can lead to the emergence of unwanted artifacts.

\subsubsection{Ablation Study on Hypergraph Construction}

\begin{figure}[tb]
    \centering
    \includegraphics[width=\linewidth]{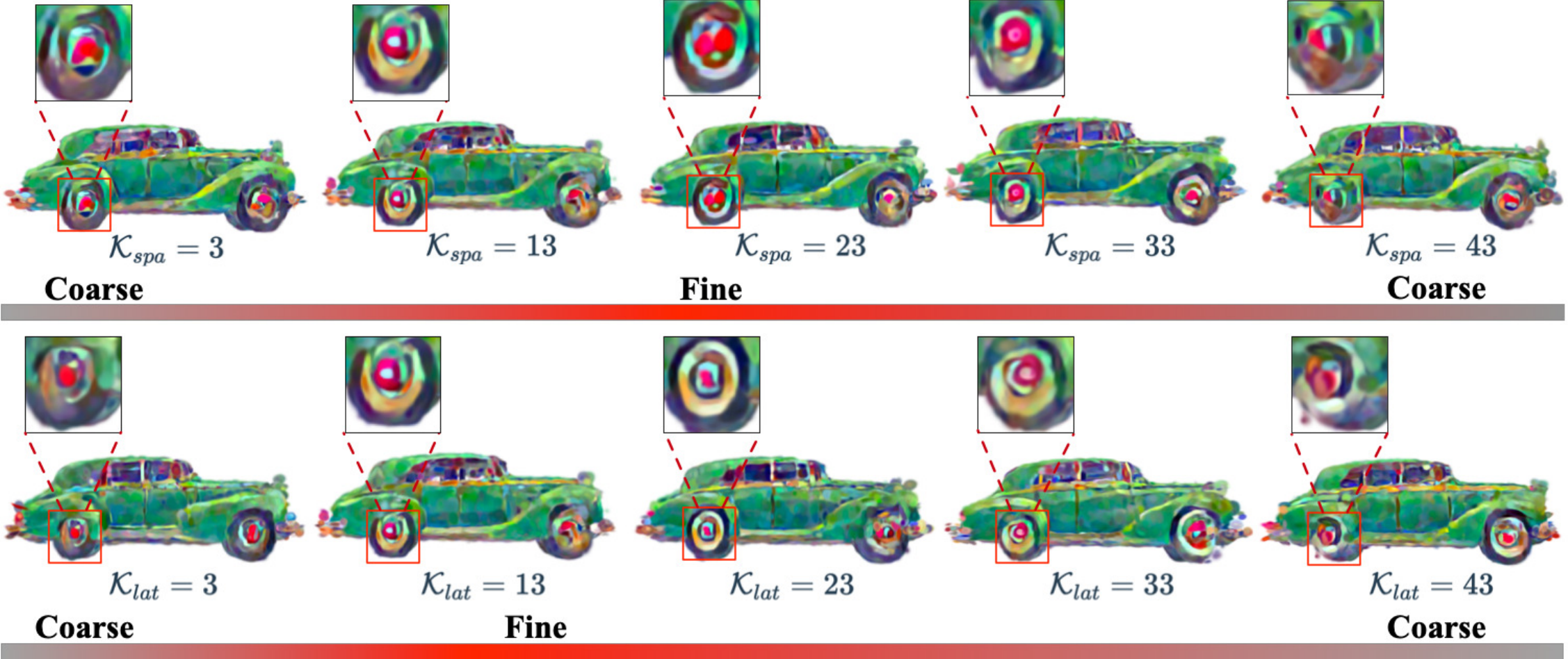}
    \caption{\textbf{Hypergraph Construction}. The comparative results of employing different hyper-parameters of constructing hypergraphs (\ie, $\mathcal{K}_{spa}$ and $\mathcal{K}_{lat}$) within our proposed framework, with identical settings maintained across all experiments.}
    \label{fig:k_lat_spa}
\end{figure}

To evaluate the effectiveness of various hyperparameters of KNN within our Hyper-3DG framework, the results of the ablation study are presented in Fig.~\ref{fig:k_lat_spa}.
Our approach utilizes two specific KNN parameters: $\mathcal{K}_{lat}$ for the image feature space and $\mathcal{K}_{spa}$ for the 3DGS parameter space.
For both KNN parameters, we conducted ablation experiments over a consistent interval with identical variables.
With $\mathcal{K}_{pat}$ set at 50, the results in Fig.~\ref{fig:k_lat_spa} suggest that the overall image rendering performance, as influenced by the KNN parameters, is comparatively favorable within a middle interval (specifically, 13-33).
For $\mathcal{K}_{lat}$, when $\mathcal{K}_{lat}=23$, the representation of rear tire is superior, while the front tire is average. 
Conversely, when $\mathcal{K}_{lat}=43$, the representation of the rear tire is markedly deficient, whereas that of the front tire surpasses the rest.
This provides guidance for parameter tuning, suggesting that a balanced representation of the tires might be achieved by targeting parameters between these two extremes.
For $\mathcal{K}_{spa}$, the representation of front tire is relatively consistent across different values of $\mathcal{K}_{spa}$, but the representation of rear tire shows a clear preference for middle values (\eg, 23) and a degradation at both extremes (\eg, 3 and 43).
The rendered image quality improves with an increase in $\mathcal{K}_{spa}$, although there is a slight decline at higher values, such as $\mathcal{K}_{spa}=43$.
These findings indicate that the optimal values for $\mathcal{K}_{lat}$ and $\mathcal{K}_{spa}$ lie within specific ranges, and they provide a basis for further refinement of the hyper-parameters to enhance the quality of the generated 3D assets.

\subsubsection{Ablation Study on Graph vs. Hypergraph}

\begin{figure}[tb]
    \centering
    \includegraphics[width=\linewidth]{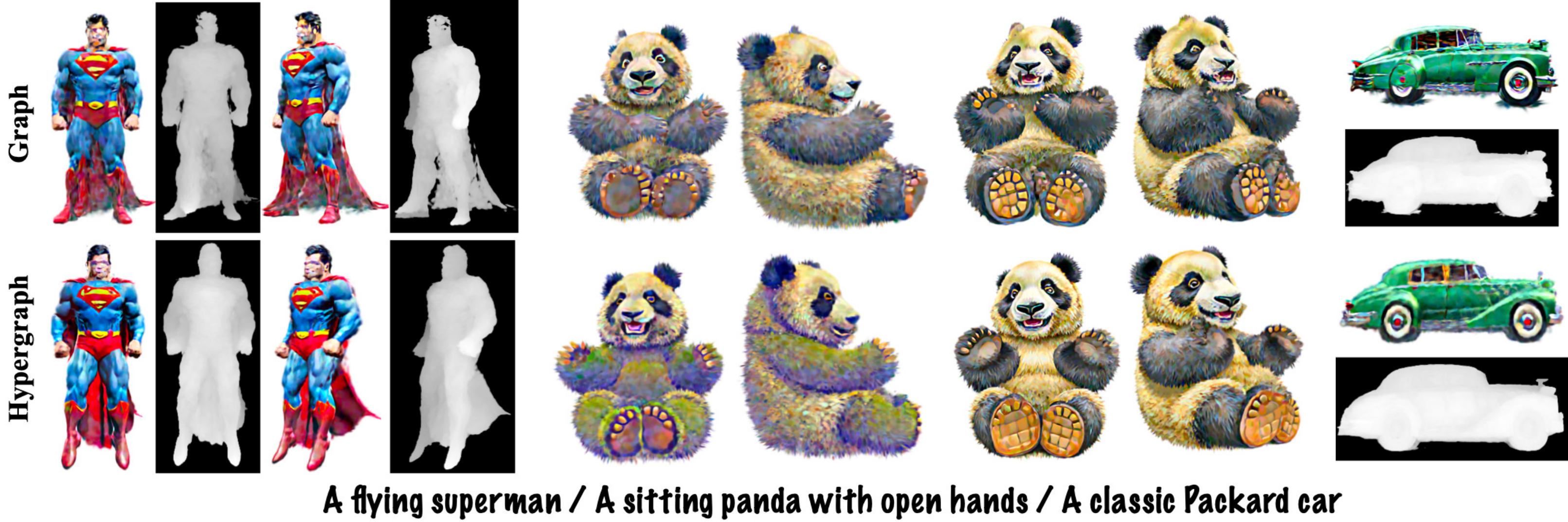}
    \caption{\textbf{Graph vs. Hypergraph}. The comparative results of employing Graph Neural Network (GNN) \cite{kipf2016semi} and our proposed hypergraph-based model model within our proposed framework, with identical settings maintained across all experiments.}
    \label{fig:g_hg_comp}
\end{figure}

We extended our comparative analysis to include graph neural networks (GNNs) \cite{kipf2016semi, velivckovic2017graph} alongside our proposed hypergraph-based methods.
In these experiments, we replaced the hypergraph convolution layer with a standard graph convolution layer (GCN) while maintaining all other settings constant.
The key difference between these two approaches lies in their ability to model relationships: the graph convolution can only capture pairwise interactions due to its inherent data structure, whereas the hypergraph convolution is capable of modeling high-order correlations among the various parts of a 3D object.
This capability is particularly advantageous for 3D data and has been widely supported by previous research in the field \cite{gao20123, gao2022hgnn+}.
As depicted in Fig.~\ref{fig:g_hg_comp}, the superiority of hypergraph-based methods is evident, such as the coherence of the flying superman's cape, the continuity of the superman's body, the symmetry between the face and body of the panda, and the overall coherence of the packard car.
These visual cues indicate that the hypergraph-based methods are more effective in processing and generating 3D data.
This result is consistent with the theoretical advantages of hypergraphs in capturing high-order correlations and complex relationships within 3D data, which is a critical aspect for achieving more realistic and detailed 3D object representations.

\subsubsection{Ablation Study on Steps of Warm Up and High-Order Refine}

\begin{figure}[tb]
    \centering
    \includegraphics[width=\linewidth]{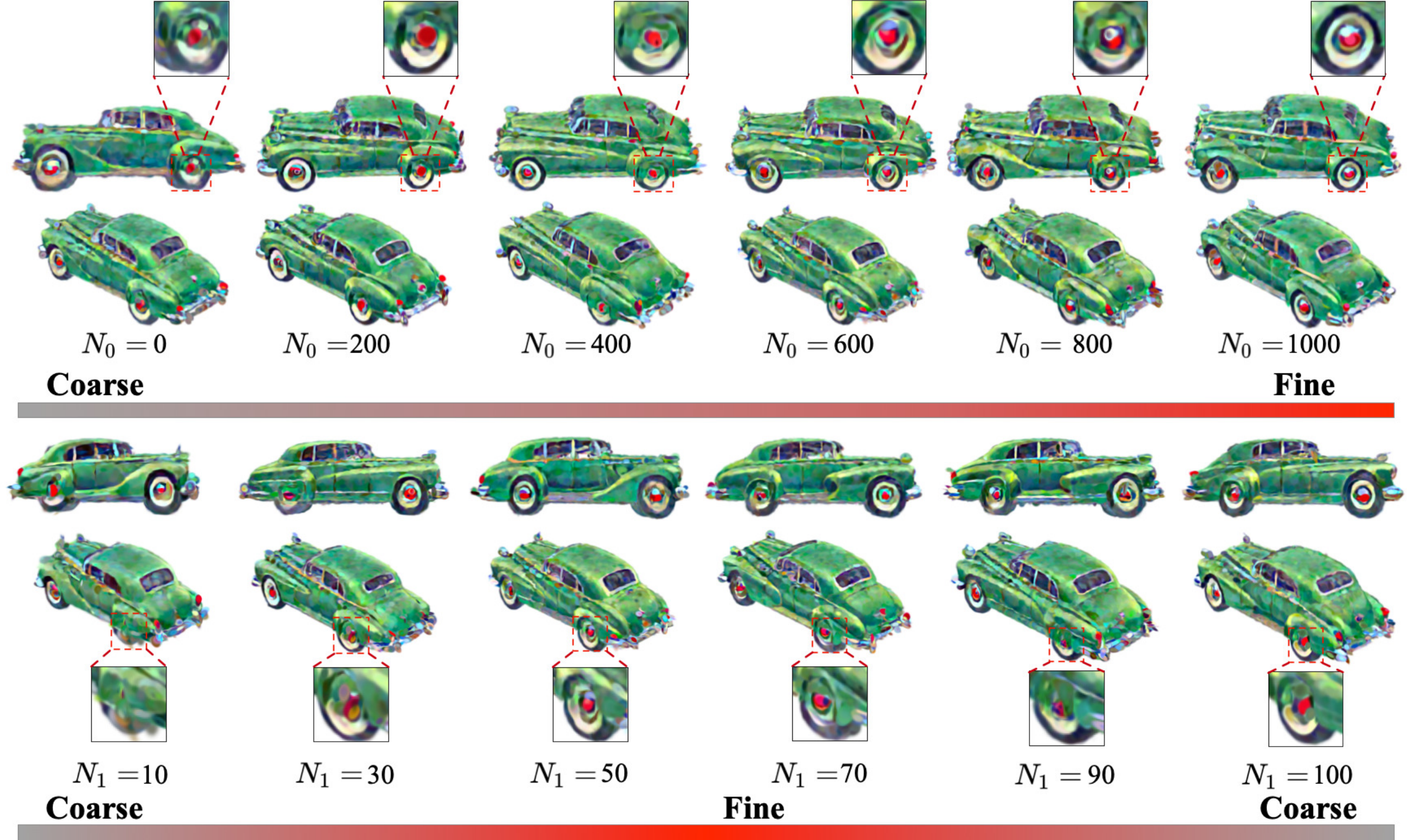}
    \caption{\textbf{Steps of Warm Up and Refine}. The comparative results of employing different hyper-parameters of ``Warm Up'' and ``Refine'' (\ie, $N_0$ and $N_1$) within our proposed framework, with identical settings maintained across all experiments.}
    \label{fig:n0_n1_fig}
\end{figure}

In our proposed framework Hyper-3DG, there are two distinct stages: ``Mainflow'' and ``High-Order Refine''.
Each of these stages is governed by two control parameters, $N_0$ and $N_1$, which we have investigated experimentally to understand their respective impacts.
As depicted in Fig.~\ref{fig:n0_n1_fig}, the impact of the initial warmup phase was investigated by varying $N_0$ from 0 to 1,000.
Notably, an $N_0$ value of 0 indicates no warmup phase, with the generation process starting directly from the initialization provided by Point-E.
The quality of the generated Packard car improved incrementally with the increase in $N_0$.
This suggests that a longer warmup phase leads to better initialization and preparation for the subsequent refinement steps.
In the refinement phase following the warmup, we noticed that the quality of the output continued to improve as $N_1$ increased from 10 to 70.
However, beyond a certain point, \ie, $N_1$ exceeding 70 and increasing to 100, the quality began to decline.
This indicates that there is an optimal range for $N_1$ that balances the refinement process without overfitting or introducing artifacts.
These findings highlight the importance of carefully selecting these hyper-parameters to achieve the best balance between computational efficiency and the quality of the generated 3D assets.

\subsubsection{Ablation Study on Pre-trained 3D Generator}

\begin{figure}[t]
    \centering
    \includegraphics[width=\linewidth]{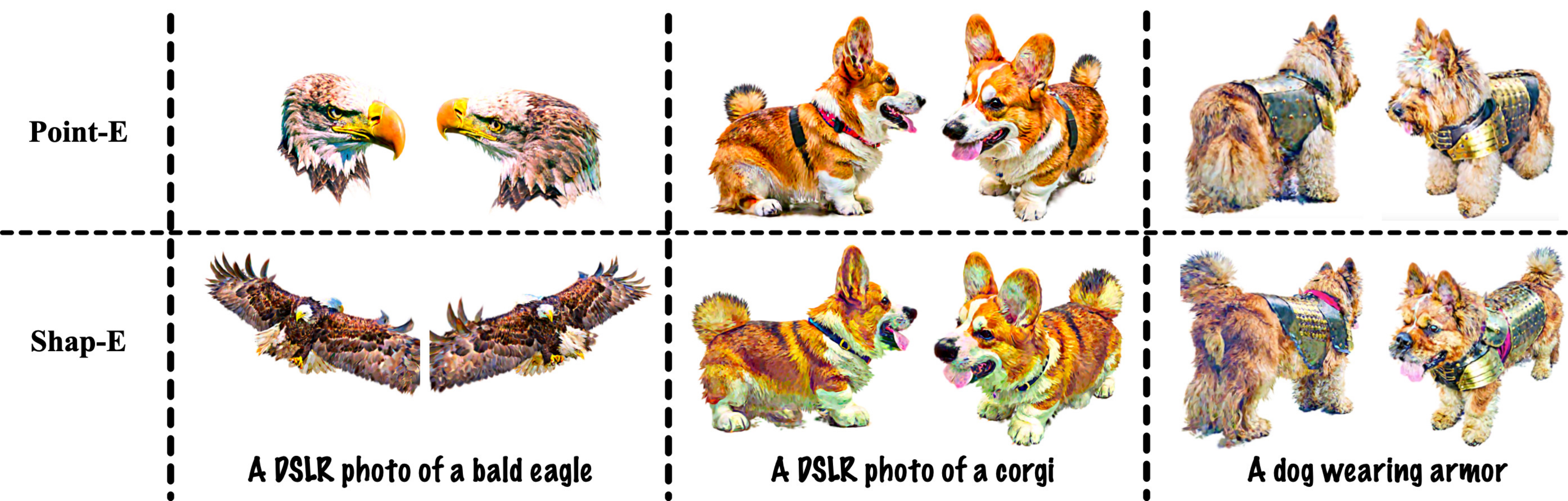}
    \caption{
    \textbf{Pre-trained 3D Generator}. The comparative results of employing different pre-trained 3D generator models (\ie, Point-E \cite{nichol2022point} and Shap-E \cite{jun2023shap}) within our proposed framework, with identical settings maintained across all experiments.}
    \label{fig:3dgen_fig}
\end{figure}

We conduct an empirical comparison to assess the impact of different pre-trained 3D generator models used for initialization, recognizing the sensitivity of 3D Gaussians to initial conditions. 
The models we compared are as follows:

\begin{itemize}
\item \textbf{Point-E} \cite{nichol2022point} is a diffusion model tailored for rapid point cloud generation. It features a transformer architecture and is capable of generating point clouds in response to text or image inputs. 
\item \textbf{Shap-E} \cite{jun2023shap} is designed to generate parameters for implicit 3D representations, such as NeRF \cite{mildenhall2020nerf} or DMTet \cite{shen2021deep}. By modeling a higher-dimensional multi-representation output space, Shap-E can quickly produce high-quality 3D assets. 
\end{itemize}

As depicted in Fig.~\ref{fig:3dgen_fig}, the bald eagle, corgi, and a dog wearing armor, all of which indicate that samples initialized with Point-E \cite{nichol2022point} are more aligned with the prompt and display fewer rough surface textures compared to those initialized using Shap-E \cite{jun2023shap}.
This finding suggests that Point-E \cite{nichol2022point} offers a more stable and precise starting point for the text-to-3D Gaussian Splatting generation process, contributing to superior overall results.

\subsubsection{Ablation Study on 2D Images Visual Feature Extractor}

\begin{figure}[t]
    \centering
    \includegraphics[width=\linewidth]{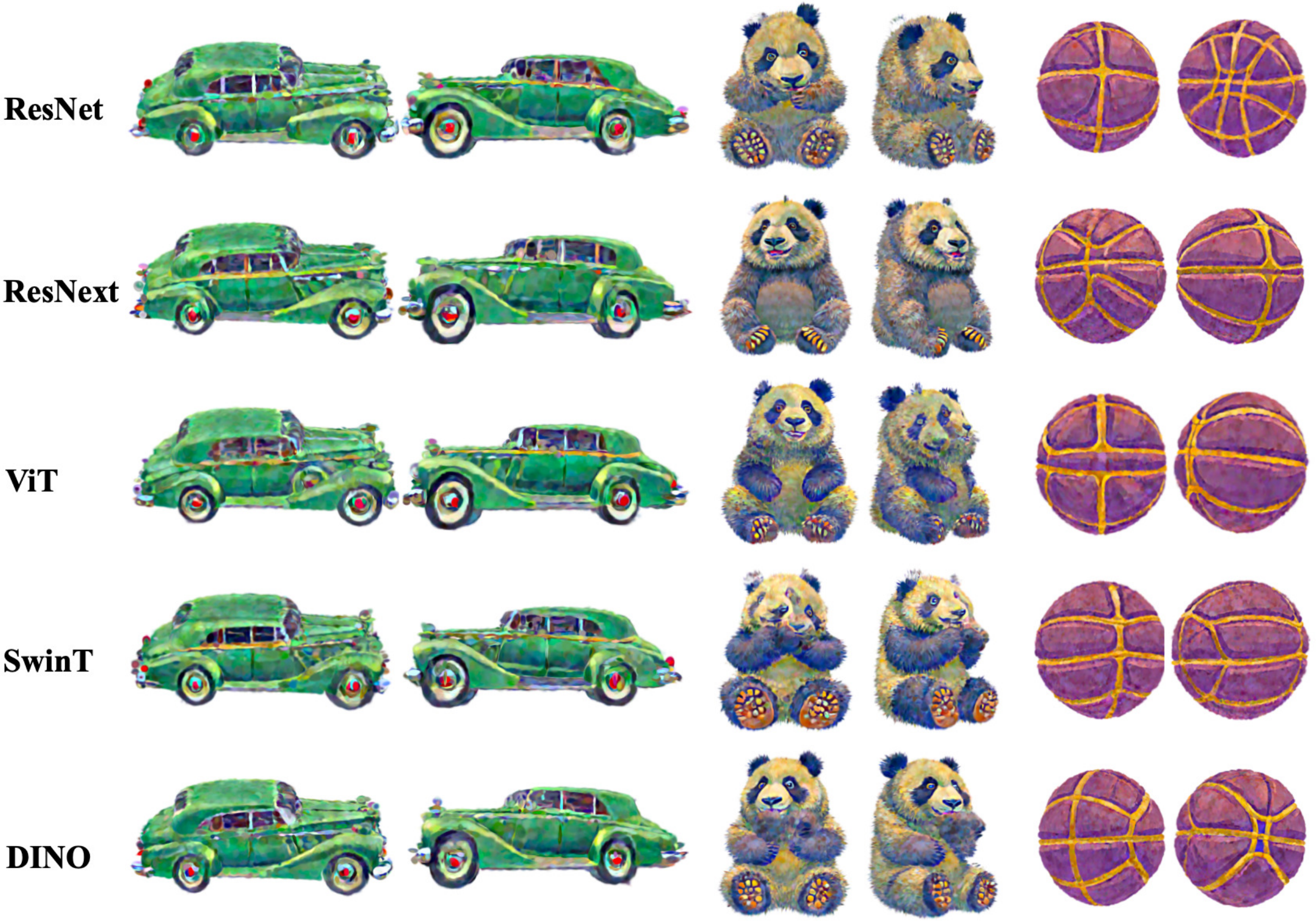}
    \caption{\textbf{Pre-trained 2D Images Visual Feature Extractor}. The comparative results of employing different pre-trained 2D Images visual feature extractor (\ie, ResNet \cite{he2016deep}, ResNeXt \cite{xie2017aggregated}, ViT \cite{dosovitskiy2020image}, Swin-T \cite{liu2021swin}, DINO \cite{caron2021emerging}) within our proposed framework, with identical settings maintained across all experiments. The new prompts utilized here encompass ``A sitting panda'' and ``A basketball''.}
    \label{fig:2dgen_fig}
\end{figure}

Our ablation experiments aimed to assess the performance of various visual feature extractors for the rendered patches.
We examined a range of models, each with its distinct characteristics and strengths in image processing.
\begin{itemize}
\item \textbf{ResNet} \cite{he2016deep} is a foundational deep residual network known for its effectiveness in addressing the vanishing gradient problem and stabilizing training, making it a standard in computer vision.
\item \textbf{ResNext} \cite{xie2017aggregated} builds upon ResNet by introducing a multi-branch architecture, which typically enhances performance in certain computer vision tasks compared to ResNet.
\item \textbf{ViT} \cite{dosovitskiy2020image} is a transformer-based architecture designed for image recognition tasks. It differs from traditional CNNs by using attention mechanisms to capture global dependencies in image data, often yielding superior results.
\item \textbf{SwinT} \cite{liu2021swin} adapts the transformer approach to computer vision with a sliding-window scheme, extending the transformer architecture to general image recognition tasks and outperforming the original ViT in many cases.
\item \textbf{DINO} \cite{caron2021emerging} is a self-supervised learning approach that leverages the ViT architecture to capture the visual semantics of images effectively, without the need for large-scale labeled datasets.
\end{itemize}
Our empirical analysis reveal that different feature extractors produce samples of varying qualities.
Samples generated with DINO \cite{caron2021emerging} were generally more detailed in terms of texture, as observed in the shape and texture of the tire of car and the color of the panda in Fig.\ref{fig:2dgen_fig}.
However, the differences among the feature extractors were not overwhelmingly significant.
Considering the balance between time and computational resources, we typically opt for ResNet or ViT as our implementation method for this part of the framework.

\subsubsection{Ablation Study on Random Render Views}

\begin{figure}[t]
    \centering
    \includegraphics[width=\linewidth]{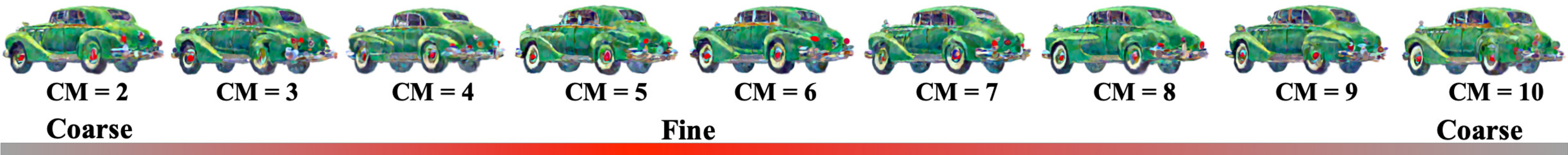}
    \caption{\textbf{Random Render Views}. The comparative results of employing different hyper-parameters of the random render views (denoted as $\textrm{CM}$) within our proposed framework, with identical settings maintained across all experiments.
    }
    \label{fig_cm}
\end{figure}

The ablation study on Random Render Views explores the effect of varying camera angles on the intermediate rendering process.
Using the prompt ``a classic Packard car'', as shown in Fig.~\ref{fig_cm}, we observe the progression of the generated quality.
Initially, at lower Camera Model (CM) values (\eg, 2 to 3), the quality of the Packard car is suboptimal, with notable deficiencies in the clarity of the tires.
As the CM values increase to the range of 4 to 7, there is a significant improvement in the generation quality, indicating that the camera angle plays a crucial role in the quality of the rendered 3D object.
However, further increasing CM values to 8 to 10 do not lead to an improvement in the quality but rather a decline, suggesting that there is an optimal range for camera angles that maximizes the visual fidelity of the generated 3D assets.

\subsection{User Study}

In the absence of standardized evaluation metrics for 3D generation, we conduct a user-centric assessment to gauge model performance.
A dedicated evaluation set was constructed, encompassing 30 prompts across five distinct methodologies.
Participants were presented with a rendered video of a particular 3D asset alongside its corresponding input text prompt.
Fifty participants independently assessed each item in the set. Evaluations focused on the asset’s alignment with the prompt and the quality of the generated details, employing a scoring system ranging from 1 to 5. 
The average scores of DreamFusion \cite{poole2022dreamfusion}, DreamGaussian \cite{tang2023dreamgaussian}, GSGEN \cite{chen2023text}, LucidDreamer \cite{liang2023luciddreamer}, and our proposed method Hyper-3DG are \textit{2.3, 2.6, 2.9, 3.6, 4.1}, respectively.
These results highlight the marked superiority of our proposed method.

\subsection{Limitations and Broader Impact}

In this section, we explore the limitations and broader implications associated with our proposed Hyper-3DG method.
The Hyper-3DG approach may yield less than optimal outcomes when faced with text prompts that contain complex scene descriptions or intricate logical structures.
This shortcoming stems from the limited language comprehension abilities of the Point-E \cite{nichol2022point} and the CLIP text encoder \cite{radford2021learning} integrated within the StableDiffusion framework.
Furthermore, although our introduced 3DGS hypergraph refiner, which incorporates 3D priors, mitigates the Janus problem, it does not entirely obviate the risk of degeneration, particularly when the textual prompt significantly influences the diffusion models.
Beyond technical challenges, the content produced by generative models could have negative implications for the labor market.
Moreover, like other generative systems, there is a risk that our method could be exploited to generate fraudulent or harmful content, underscoring the need for heightened vigilance and ethical considerations in its application.
\section{Conclusion and Future Work}

In summary, our research introduces Hyper-3DG, a framework that seamlessly integrates differentiable rendering and text-to-image advancements to efficiently generate high-quality 3D assets.
Central to our approach is the Geometry and Texture Hypergraph Refiner (HGRefiner), which effectively overcomes the Janus problem and the inherent incoherence issue in generation processes.
Hyper-3DG can be applied to various differentiable 3D representations, generally enhancing the quality and reducing the time consumption of existing 3D generation methods.
This work not only advances the quality and diversity of 3D assets but also sets a precedent for future innovations in 3D modeling, with far-reaching implications for virtual reality and gaming industries.
In future work, we will focus on generating more sophisticated 3D objects as well as intricate scenes together by improving the ability to leverage the capabilities of pre-trained 2D and 3D generation models.
\section{Declarations}

\begin{itemize}
    \item Data Availability Statement:
    The data used in this study are not publicly available due to the nature of the generative task. The research presented in this paper is based on simulated data and does not involve the collection or usage of any public or private datasets. The simulation data were generated internally for the purpose of this study and are not accessible to external researchers. However, the methods and findings presented in this paper are replicable, and the authors are willing to share the code used for data generation with interested researchers upon reasonable request. 
    \item Code Availability Statement:
    The generated results reported in this paper are available in the public repository (\url{https://github.com/yjhboy/Hyper3DG}). Additionally, the code used to generate these results will be released in the same repository soon. The repository will be accessible to the research community, allowing for reproducibility of the experiments and further exploration of the methods presented in this study.
    \item Conflict of Interest Statement:
    The authors of this research paper declare that there are no conflicts of interest regarding the content of this study. None of the authors have any financial, personal, or professional affiliations that could be perceived as having influenced their work on this paper.
    \item Compliance with Ethical Standards Statement:
    This research study was conducted in accordance with the ethical standards set forth by the relevant institutional review board. All procedures involving human participants were approved by the board, and all participants provided informed consent prior to participating in the study. The authors ensure that all data used in this study are anonymized and treated with confidentiality.
    \item Informed Consent Statement:
    All participants in this study provided informed consent before participating in the research. They were fully informed about the purpose of the study, the procedures involved, and the potential risks and benefits. Participants were also informed about their right to withdraw from the study at any time without consequences. The informed consent forms were signed by the participants and are kept securely by the researchers.
\end{itemize}

\backmatter

\bibliography{sn-bibliography}

\end{document}